\documentclass{article}
\usepackage{ijcai24}

\usepackage{times}
\usepackage{soul}
\usepackage{url}
\usepackage[hidelinks]{hyperref}
\usepackage[utf8]{inputenc}
\usepackage[small]{caption}
\usepackage{graphicx}
\usepackage{amsmath}
\usepackage{amsthm}
\newtheorem{definition}{Definition}
\usepackage{amsfonts}
\usepackage{booktabs}
\usepackage{algorithm}
\usepackage{algorithmic}
\usepackage[switch]{lineno}
\usepackage[utf8]{inputenc}
\usepackage{tikz}
\usepackage{forest}
\usetikzlibrary{trees,positioning,shapes,shadows,arrows.meta}

\title{Graph Domain Adaptation: Challenges, Progress and Prospects}

\author{
Boshen Shi$^{1,2}$
\and
Yongqing Wang$^1$\textsuperscript{*}\and
Fangda Guo$^1$\textsuperscript{*}\and
Bingbing Xu$^1$\and \\
Huawei Shen$^{1,2}$\and
Xueqi Cheng$^{1,2}$
\affiliations
$^1$CAS Key Laboratory of AI Safety\&Security, Institute of Computing Technology, CAS\\
$^2$University of Chinese Academy of Sciences\\
\emails
\{shiboshen19s, wangyongqing, guofangda, xubingbing, shenhuawei, cxq\}@ict.ac.cn
}

\begin{document}

\maketitle

\begin{abstract}
As graph representation learning often suffers from label scarcity problems in real-world applications, researchers have proposed graph domain adaptation (GDA) as an effective knowledge-transfer paradigm across graphs. In particular, to enhance model performance on target graphs with specific tasks, GDA introduces a bunch of task-related graphs as source graphs and adapts the knowledge learnt from source graphs to the target graphs. Since GDA combines the advantages of graph representation learning and domain adaptation, it has become a promising direction of transfer learning on graphs and has attracted an increasing amount of research interest in recent years. In this paper, we comprehensively overview the studies of GDA and present a detailed survey of recent advances. Specifically, we outline the research status and challenges, propose a taxonomy, introduce the details of representative works, and discuss the prospects. To the best of our knowledge, this paper is the first survey for graph domain adaptation. A detailed paper list is available at https://github.com/Skyorca/Awesome-Graph-Domain-Adaptation-Papers.
\end{abstract}

\renewcommand{\thefootnote}{\fnsymbol{footnote}}
\footnotetext[1]{Corresponding authors}

\section{Introduction}
Recently, graph representation learning has become prevalent in real-world applications, with the aim of generating meaningful embeddings through exploring entity semantics and complicated pairwise relationships. However, developing effective graph models for specific tasks typically relies on supervised training, which may suffer from label scarcity problems as annotating complex structured data is both expensive and difficult. This challenge necessitates transferring abundant labeling knowledge from task-related graphs. Given that these auxiliary graphs (source) and the task graphs (target) may exhibit diverse distributions in node attributes, structural patterns, and task labels, which could hinder knowledge transfer, researchers have proposed graph domain adaptation (GDA) as a paradigm to effectively transfer knowledge across graphs by addressing these intricate distribution shifts.

Generally, pioneering works establish cross-network classification (aka cross-domain classification) tasks for transferring knowledge on graphs. They straightforwardly employ deep domain adaptation (DA) techniques, typically applied to unstructured data, to learn domain-invariant and label-discriminative embeddings~\cite{CDNE,ACDNE,ADAGCN,ASN}. These embeddings generated by shared graph encoders range from node-level to edge-level and up to graph-level, depending on the granularity of tasks. While early studies often overlook complex graph structures and distribution shifts in topological attributes, follow-up works have progressed by delving into topologies and diverse structural shifts~\cite{GraphAE,SpecReg,mywork2}. For example, GraphAE~\cite{GraphAE} studies how shifts in node degree distribution contribute to domain discrepancy and eliminates such discrepancy via aligning message routers. In addition, researchers have explored other distribution shifts, such as conditional shifts, and proposed more strategies to facilitate better knowledge transfer on graphs.
Expanding beyond theoretical models, GDA finds practical applications in areas such as anomaly detection~\cite{ACT} and urban computing~\cite{CrossTReS}. Moreover, various evaluation benchmarks have been established for GDA models, such as OpenGDA~\cite{mywork3}. 

Transferring knowledge on graphs is a burgeoning area with two main paradigms: GDA and graph pre-training\&fine-tuning/prompt-tuning. Graph pre-training has proven effective in transferring general structure-relevant knowledge from vast unlabeled graphs, but it requires few labels in downstream task graphs. In contrast, GDA directly transfers task-specific knowledge from labeled source graphs and remains practical even with unlabeled task graphs. Therefore, GDA has demonstrated its unique value in graph transfer learning, necessitating further research on models, theoretical foundations, and applications. We consider this an opportune time to present a comprehensive survey that aids researchers in understanding the fundamental topics and key directions in the field. To the best of our knowledge, this survey is the first to provide an overview and future perspective of GDA.

The rest of this survey is organized as follows: We begin with a detailed formulation of GDA, following which we analyze the key challenges in the field and introduce a novel taxonomy that categorizes existing literature from three perspectives: source-based, adaptation-based, and target-based. We also explore various extensions and applications. Finally, we discuss potential avenues for future research, offering insights that could further advance this promising field.

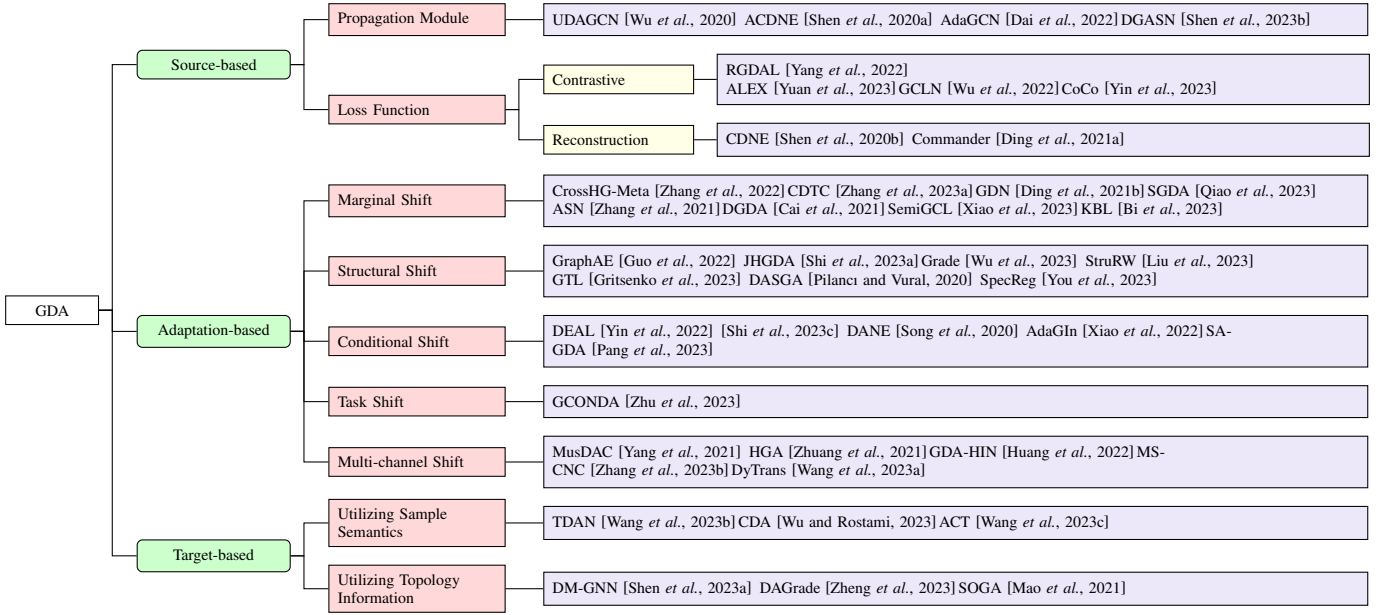
\begin{figure*}
    \centering

\tikzset{
    basic/.style  = {draw, text width=1cm, align=center, font=\tiny, rectangle},
    root/.style   = {basic, rounded corners=2pt, thin, align=center, fill=green!30},
    onode/.style = {basic, thin, rounded corners=2pt, align=center, fill=green!60,text width=3cm,},
    tnode/.style = {basic, thin, align=left, fill=pink!60, text width=6em},
    xnode/.style = {basic, thin, rounded corners=2pt, align=center, fill=green!20,text width=1.8cm,},
    wnode/.style = {basic, thin, align=left, fill=pink!10!blue!80!red!10, text width=30.5em},
    w2node/.style = {basic, thin, align=left, fill=pink!10!blue!80!red!10, text width=24em},
    snode/.style = {basic, thin, align=left, fill=pink!10!yellow!10, text width=5em},
    edge from parent/.style={draw=black, edge from parent fork right}

}

\begin{forest} for tree={
    grow=east,
    growth parent anchor=east,
    parent anchor=east,
    child anchor=west,
    anchor=center, 
    edge path={\noexpand\path[\forestoption{edge},-, >={latex}] 
         (!u.parent anchor) -- +(5pt,0pt) |-  (.child anchor) 
         \forestoption{edge label};}
}
[GDA, basic,  l sep=5mm,
    [Target-based, xnode,  l sep=5mm,
      [Utilizing Topology \\Information, tnode, l sep=5mm, 
        [DM-GNN~\cite{DM-GNN}\, DAGrade~\cite{DAGrade}\,SOGA~\cite{SFUGDA} ,wnode]]
        [Utilizing Sample \\Semantics, tnode, l sep=5mm,[TDAN~\cite{TDAN}\,CDA~\cite{CDA}\,ACT~\cite{ACT},wnode]]
    ]
    [Adaptation-based, xnode, l sep=5mm,
        [Multi-channel Shift, tnode, l sep=5mm, [MusDAC~\cite{MusDAC}\, HGA~\cite{HGA}\,GDA-HIN~\cite{GDA-HIN}\,MS-CNC~\cite{MS-CNC}\,DyTrans~\cite{DyTrans}, wnode]]
        [Task Shift, tnode, l sep=5mm, [GCONDA~\cite{GCONDA},wnode]]
        [Conditional Shift, tnode, l sep=5mm, [DEAL~\cite{DEAL}\,~\cite{mywork1}\, DANE~\cite{DANE2}\, AdaGIn~\cite{AdaGIn}\,SA-GDA~\cite{SA-GDA},wnode]]
        [Structural Shift,tnode, l sep=5mm,[GraphAE~\cite{GraphAE}\, JHGDA~\cite{mywork2}\,Grade~\cite{GRADE}\,  StruRW~\cite{StruRW}\, GTL~\cite{GTL}\, DASGA~\cite{SDA2}\, SpecReg~\cite{SpecReg},wnode]]
        [Marginal Shift,tnode, l sep=5mm, [CrossHG-Meta~\cite{CrossHG-Meta}\,CDTC~\cite{cdtc}\,GDN~\cite{GDN}\,SGDA~\cite{SGDA}\,\\ASN~\cite{ASN}\,DGDA~\cite{DGDA}\,SemiGCL~\cite{SemiGCL}\,KBL~\cite{KBL},wnode]]
    ]
    [Source-based, xnode,  l sep=5mm,
        [Loss Function,tnode, l sep=5mm,
        [Reconstruction, snode, l sep=3mm,[CDNE~\cite{CDNE}\,  Commander~\cite{COMMANDER},w2node]]
        [Contrastive, snode, l sep=3mm,[
        RGDAL~\cite{RGDAL}\, ALEX~\cite{Alex}\,GCLN~\cite{GCLN}\,CoCo~\cite{COCO}\,
        ,w2node]]
        ] 
        [Propagation Module, tnode,l sep=5mm,
        [UDAGCN~\cite{UDAGCN}\, ACDNE~\cite{ACDNE}\,  AdaGCN~\cite{ADAGCN}\,DGASN~\cite{CNHHEC}  , wnode]]
    ] 
]
\end{forest}

\caption{Taxonomy of Graph Domain Adaptation.}
\label{fig:lit_surv}
\end{figure*}

\section{Problem Formulation}
\label{sect::formulation}
Following definitions in transfer learning and previous formulations of GDA~\cite{survey1,survey2,ADAGCN,GraphAE,StruRW}, we introduce a more comprehensive and detailed formulation.
\begin{definition}[Graph Domain]
A Graph $G$ consists of node feature matrix $X\in \mathbb{R}^{n\times h}$ and adjacency matrix $A\in \mathbb{R}^{n\times n}$, where $n$ indicates size of node set $\vert V\vert$ and $h$ is raw feature dimension. Therefore, $G=(X,A)$, and a graph domain $\mathcal{D}$ equals to $\{(G_i,P(G_i))\vert_{i=1}^m\}$ where $m>1$ indicates a domain containing multiple graphs. $P(G)$ includes distributions of node feature $P(X)$ and topological attributes $P(A)$.
\end{definition}

\begin{definition}[Task]
Given a graph domain $\mathcal{D}$, a task $\mathcal{T}$ on $\mathcal{D}$ consists of a predictive function $f(\mathcal{D})$ and label space $\mathcal{Y}$, i.e.,  $\mathcal{T}=\{f(\mathcal{D}),\mathcal{Y}\}$, which are associated with node, edge or graph depending on the granularity of tasks. Each task could have groundtruth information $Y$.
\end{definition}
\begin{definition}[Graph Domain Adaptation]
    Given $m^S$ source domain(s) and task(s) (i.e. $\{(\mathcal{D}^{S_i},\mathcal{T}^{S_i})\vert_{i=1}^{m^S}\})$, and $m^T$ target domain(s) and task(s) (i.e. $\{(\mathcal{D}^{T_j},\mathcal{T}^{T_j})\vert_{j=1}^{m^T}\})$, GDA adapts one or more source domains with richer groundtruth information to transfer knowledge and improve the performance of the target learner $f^{T_j}$ on target domains. 
\end{definition}

Typically, the source and target domains should be explicitly or implicitly related, such as different online social platforms~\cite{ACT} or proteins from various species~\cite{SpecReg}. Thus, GDA often requires that the node feature space $\mathcal{X}$ and the task $\mathcal{T}$ are consistent across all domains.   Despite the domain relevance, there exist significant domain shifts, such as marginal distribution shift ($P(X^{S_i})\neq P(X^{T_j})$). We further introduce some other domain shifts that have been studied in GDA literature without violating the original definition. 1) \textit{Conditional shift}: we refer to $P(X^{S_i}\vert Y^{S_i})\neq P(X^{T_j}\vert Y^{T_j})$ as conditional shift, which indicates the label-conditional distribution of node features shifts across domains. 2) \textit{Task shift}: we refer to $P(Y^{S_i}\vert X^{S_i})\neq P(Y^{T_j}\vert X^{T_j})$ as task shift, which indicates the groundtruth predictive function $p(y|x)$ could vary across domains. 3) \textit{Structural shift}: we use $P(A^{S_i})\neq P(A^{T_j})$ to indicate probability distributions of specific topological attribute shift across domains, such as node degree and subgraphs. Moreover, we use $P(A^{S_i}\vert Y^{S_i})\neq P(A^{T_j}\vert Y^{T_j})$ to represent label-conditioned structural shifts across domains. For example, as the correlations between node labels are captured by graph structure, the distribution of edges with label-label correlations varies across domains. Overall, the above domain shifts will be further discussed in the following sections with the corresponding literature.

\section{Challenges and Categorization}
We summarize the main challenges of GDA from two aspects, which mainly stem from the unique properties of graph data. 
Firstly, the non-Euclidean nature of graph data, coupled with the diversity of graph-related tasks, poses difficulties in learning effective node embeddings and reducing source prediction errors, which is crucial to adaptation as per generalization bound theory.
Secondly, adapting knowledge learnt on source domains to target domains is challenging due to various domain shifts. These shifts are not only caused by differences in feature distributions but also by complex structural variances across domains, which appear in different forms such as differences in node degree distribution, subgraph distribution, and overall scale. These structural shifts mark a significant distinction between GDA and conventional DA on unstructured data.

To tackle these challenges, considerable efforts have been made in literature, which can be categorized into three classes:
\begin{itemize}
    \item \textbf{Source-based}: Methods in this category focus on reducing source prediction risks through enhanced node embeddings. Following generalization bound theory, this approach should result in a corresponding reduction in target risk. By employing customized graph encoders with task-specific propagation modules and loss functions, node embeddings are improved in discriminability, robustness, and generalizability.
    \item \textbf{Adaptation-based}: Methods in this category contribute to bridging domain gaps by eliminating the complex graph domain shifts. An effective way involves directly reducing marginal distribution shifts across domains. While this strategy may align overall embedding distributions, it lacks characterizing of differences in complex graph structures and task-specific components, resulting in inferior results. 
    Consequently, in addition to marginal shifts, researchers have proposed methods to eliminate these structure- or task-relevant shifts across domains.
    \item \textbf{Target-based}: Methods in this category improve target embeddings and refine target predictions by leveraging information from target domains, including sample semantics and graph topology.
\end{itemize}
Therefore, we summarize existing works with the proposed taxonomy in Figure~\ref{fig:lit_surv}. Some models tailored for particular scenarios and data are not included; they are further discussed in Section~\ref{sect::ext} and Section~\ref{sect::app}.
\section{Source-based GDA}
In general, this line of GDA research can be broadly categorized from two aspects: propagation module-based and loss function-based. We introduce them in detail below.
\subsection{Propagation Module-based Methods}
Recent studies have improved propagation modules in graph encoders (e.g., GNNs) to enhance node embeddings. A typical method is utilizing global consistency on graphs, i.e., the topological proximity between nodes within $k$ steps on graphs ($k>1$). While existing works mainly propagate information between direct neighbors (local consistency), propagating between high-order neighbors brings additional merits to the aggregated embeddings. Firstly, as nodes with similar labels and patterns may reside in different subgraphs without direct connections, aggregating from distant neighbors could enhance the discriminability of node embeddings. 
In addition, explicitly learning global information contributes to more precise adaptation from a global perspective. 

Therefore, to capture high-order proximity as global consistency on graphs, UDAGCN~\cite{UDAGCN} and ACDNE~\cite{ACDNE} adopt $k$-hop PPMI matrix. Specifically, UDAGCN develops a dual-GNN which propagates information between both local neighbors and high-order neighbors to derive two-brunch embeddings, followed by an attention mechanism to combine these two views. Dual-GNN is widely adopted by follow-up studies~\cite{ASN,SGDA,SA-GDA}. Besides, ACDNE also designs feature extractors to aggregate from local and high-order neighbors. AdaGCN~\cite{ADAGCN}, on the other hand, propagates information on the $k^{th}$ power of the adjacency matrix. In addition, DGASN~\cite{CNHHEC} employs Graph Attention Network (GAT) and applies direct supervision on attention coefficients based on the observed edge labels (i.e., homophilous or heterophilous) from source domains. Thus, it lowers the impact of heterophilous edges in propagation, enhancing the discriminability of embeddings.

\subsection{Loss Function-based Methods}
\subsubsection{Contrastive Loss} 
Contrastive learning is adopted for GDA models to enhance the quality of embeddings, especially in source domains. 1) \textit{Enhancing Robustness}:  Real-world graphs often contain structural noise and incorrect labels, making learning and transferring from them potentially sub-optimal and unreliable. RGDAL~\cite{RGDAL} addresses this problem by employing graph mutual information between graph-level raw features and embeddings, prompting GNN encoders to ignore noisy factors during embedding. ALEX~\cite{Alex} first creates a homophily-enhanced augmented graph view using a low-rank adjacency matrix obtained through singular value decomposition. It then contrasts node embeddings between the original and augmented views to derive contrastive losses, effectively filtering out noise. 
2) \textit{Exploiting Topology}: Some studies exploit richer topological information to enhance embeddings. For example, GCLN~\cite{GCLN} contrasts local-view (node embedding) with global-view (graph embedding) for a node to learn global consistency. For cross-domain graph classification tasks, CoCo~\cite{COCO} employs a hierarchical graph kernel network to generate a hierarchical view, which is then contrasted with a much flatter view from global pooling. Thus, the contrastive loss from these complementary views substantially improves the quality of graph embeddings.

\subsubsection{Reconstruction Loss} 
Reconstruction is employed as a task to learn domain-specific information. For example, CDNE~\cite{CDNE} employs stacked autoencoders (SAEs) to learn node embeddings. By minimizing the reconstruction errors of SAEs, node embeddings capture domain-specific topological properties. For cross-domain anomaly detection, Commander~\cite{COMMANDER} utilizes GAT decoder to reconstruct raw node features from embeddings and minimizes such reconstruction loss. Since anomalies typically exhibit poor reconstruction with significant errors, using reconstruction effectively characterizes domain-specific anomaly patterns.

\section{Adaptation-based GDA}
Adapting the knowledge learnt on source graph domains to target graph domains is crucial yet challenging due to the intricate domain shifts. To address this problem, the existing GDA works have introduced innovative adaptation strategies, and they can be classified into five categories based on the domain shift they mainly address. We provide a detailed introduction to them in the following sections.

\subsection{Marginal Shift}
Marginal distribution shift is the most fundamental and critical domain shift. Thus, existing GDA works develop various strategies to address it, and a simple yet effective method is modeling and reducing it with discrepancy measurement~\cite{CDNE,CW-GCN} or adversarial training~\cite{DANE,ADAGCN}. In this section, we introduce those studies that design strategies to eliminate marginal distribution shifts by leveraging methods from other research areas.
\subsubsection{Meta-Learning for GDA}
Meta-learning, particularly MAML-based approaches, could be adopted to improve the existing GDA methods. They extract comprehensive meta-knowledge from source domains to learn a well-generalized model initialization, which could fast adapt to target domains~\cite{metaDA}. For example, CrossHG-Meta~\cite{CrossHG-Meta} combines the advantages of meta-learning and GDA by simulating source-target distributions within the inner-loop of MAML, considering the inaccessibility to target domain during meta-training. Specifically, it splits all available source domains into virtual source and target domains and eliminates marginal shifts between virtual domains with discrepancy measurement. Moreover, CDTC~\cite{cdtc} addresses source-target shifts in meta-learning by selecting the most target-relevant meta-tasks in source domains for meta-training. It requires a few labeled target samples to form target prompt tasks. On the other hand, GDN~\cite{GDN} has experimentally demonstrated that using meta-learning alone can also enhance model performance in cross-domain node classification tasks.

\subsubsection{Data Augmentation for GDA}
As the structured graph data may have more input disparity than unstructured data, only using the parameters in the graph encoders (e.g., shallow GNNs) may be insufficient to mitigate the distribution shifts~\cite{augmentDA}. Therefore, data augmentation could be used with existing GDA methods to improve adaptation performance. For example, SGDA~\cite{SGDA} augments original source graphs by adding trainable perturbations (i.e., adaptive shift parameters) to embeddings and utilizes adversarial learning to train both the graph encoder and perturbations for reducing marginal shifts. Since these perturbations are optimized for domain invariance, they align the distribution of source embeddings more closely with that of the target embeddings. Furthermore, DEAL~\cite{DEAL} advances the strategy by adding perturbations to both raw node features and embeddings, which has proven more effective than simply perturbing embeddings alone.

\subsubsection{Disentanglement for GDA}
Disentanglement-based models mainly extract the domain-invariant variables from the embedding space with entangling domain-specific information~\cite{disentangleDA}. For example, ASN~\cite{ASN} directly separates domain-specific and domain-invariant variables by designing a private encoder for each domain and a shared encoder across domains, forcing these two types of encoders to extract different features. The domain-invariant node embeddings are used for adaptation, while domain-specific ones are used to reconstruct topology. DGDA~\cite{DGDA}, on the other hand, believes it is challenging to perfectly disentangle private and shared variables for complex graph data. Instead, it disentangles three latent variables, which could control the graph generation process, from each graph: domain-specific variables, graph semantic variables, and random variables. Specifically, it employs three variational graph auto-encoders (VGAEs) to encode these variables and further disentangles them with corresponding decoders, i.e., domain classifier, graph-level classifier and noise reconstruction module. 

\subsubsection{Other Strategies}
We introduce some other adaptation strategies that have been proven effective on graphs. SemiGCL~\cite{SemiGCL} adopts minimax entropy-based methods~\cite{mmeDA}, wherein the node classifier is trained to maximize the entropy of target predictions while encoders minimize such entropy. Besides, KBL~\cite{KBL} first applies adversarial training and data augmentation to align node embeddings across domains. It then establishes a Bridged-graph in the common embedding space by connecting knowledgeable source nodes to each target node, following which the propagation module of GNNs transfers sample-wise knowledge across domains. GCLN~\cite{GCLN} alternatively reduces marginal shifts with contrastive loss. Specifically, it encourages the similarity between node embeddings in the source (target) domain and the graph-level summary embedding in the target (source) domain.

\subsection{Structural Shift}
The intrinsic properties of graph structures lead to structural domain shifts and bring additional challenges to GDA. Consequently, researchers have proposed methods to alleviate the structural shifts from both spatial and spectral domains.

\subsubsection{Spatial methods} 
\noindent\textbf{Node degree}: GraphAE~\cite{GraphAE} studies how shifts in node degree distribution affect node embeddings. It observes that given source and target graphs with different densities (i.e., different degree distributions), messages spread more quickly in the denser graph, leading to a more comprehensive exploration of global topology and, consequently, differences in node embeddings. Based on this insight, it develops a message router which learns router embeddings for each node, indicating how much information is propagated to it. Since nodes with higher degrees receive more messages, and vice versa, their router embeddings could encode the degree information. Consequently, GraphAE minimizes the discrepancy between router embedding distributions to eliminate structural shifts, in addition to other adaptation losses. 

\noindent\textbf{Network hierarchy}: JHGDA~\cite{mywork2} studies the shifts in hierarchical graph structures, which are inherent properties of graphs. Such shifts lead to distinct node embeddings across domains. An example comes from online social networks (OSNs), where a user could be influenced by other users distributed in different communities, and the distributions of communities, including their concepts, quantities and scales, are totally different across OSNs. To characterize hierarchical structure shifts, JHGDA first learns a hierarchical pooling model to extract meaningful hierarchies for source and target domains. It then aggregates domain discrepancy from all hierarchy levels to derive a comprehensive discrepancy measurement, which has been proven to be the combination of differentiable graph kernels. 

\noindent\textbf{Subtree}: Grade~\cite{GRADE} captures shifts in subtree distributions motivated by the connection of non-parametric Weisfeiler-Lehman (WL) graph kernel and parametric GNNs, wherein the former decomposes graphs into a sequence of subtrees rooted at every node, and the latter iteratively aggregates messages from nodes' local neighborhoods. Specifically, it proposes a novel graph subtree discrepancy by aggregating distribution shifts of subtrees from all depths. The subtree representation at depth $l$ of any node is obtained from the updated node embedding after the $l^{th}$ aggregation. 

In addition to studying the distribution shifts of particular topological attributes, StruRW~\cite{StruRW} addresses the conditional structural shifts across domains in node-level tasks. As depicted in Section~\ref{sect::formulation}, such shifts indicate that connection patterns between classes vary across domains, which significantly increases node embedding shifts after GNN propagation. Therefore, StruRW re-weights each source edge to ensure the probabilities of connecting a class-$i$ node to class-$j$ node remain the same across domains for any classes $i$ and $j$. Moreover, GTL~\cite{GTL} alleviates structural shifts by learning an isomorphism mapping to match source and target graphs. Although GTL could transfer structural node labels across domains, applying it to real-world graphs is challenging as the isomorphism assumptions may not hold and scaling to large graphs is under-explored.

\subsubsection{Spectral methods} 
By formulating predictive functions $p(y|x)$ as graph signals, DASGA~\cite{SDA2} discovers that the shifts in the inherent spectral properties across graph domains lead to potential task shifts. To overcome these problems, it learns and transfers the spectral content of the predictive function from source to target domain. Specifically, it aligns graph Fourier basis across domains with learning a source-to-target mapping. It additionally proves that the smoothness of the target predictive function is upper-bounded, which guarantees the effectiveness of adaptation to a certain degree. Besides, SpecReg~\cite{SpecReg} re-formulates the optimal transport (OT)-based DA bound for graph data and discovers that adjusting the GNNs' Lipschitz constant can tighten the bound. Such constant is strongly related to inherent structural differences between domains. Therefore, in addition to optimizing the source prediction errors and the domain divergence terms, it regulates GNN spectral properties, including spectral smoothness (SS) and maximum frequency response (MFR) on both source and target domains. To the best of our knowledge, it proposes the first model-agnostic generalization bound for GDA.

\subsection{Conditional Shift}
\label{sect::cond_shift}
For cross-network classification tasks, reducing conditional shifts across domains enhances the discriminability of target embeddings. It typically involves two necessary stages, i.e., generating pseudo-labels on unlabeled target domains and aligning distributions of label-conditioned embeddings across domains. Thus, the existing works considering conditional shifts broadly fall into two categories based on which stage they primarily focus on.

\subsubsection{Generating pseudo-labels} GDA models always need to address the problems with limited or even no target labels. Thus, it is critical to generate reliable pseudo labels on target domains for calculating conditional distributions. A straightforward solution directly uses target predictions from source classifiers as pseudo labels~\cite{CDNE,DM-GNN,Alex}. However, when the source and target distributions are not well-aligned and the domain gap is non-negligible, these pseudo-labels could be overconfident on target domains and noisy, resulting in potential biases during optimization and thus influencing the performance~\cite{DEAL,CDA}. 

DEAL~\cite{DEAL}, for example, believes that consistent target predictions from different views are more likely to be reliable. As deep and shallow layers of GNN encoders focus on different perspectives, i.e., global and local topologies, DEAL filters out those noisy nodes whose pseudo labels are inconsistent between these two complementary views. In practice, pseudo labels from deep layers are generated by source classifiers, while those from shallow layers are computed by clustering algorithms like DBSCAN. Besides,~\cite{mywork1} combines node semantic information and graph topology to obtain high-quality pseudo labels. In particular, target nodes are first labeled by their nearest source class centres in the
embedding space. These pseudo-labels are further propagated on target networks for smoothing because of network homophily (i.e., neighboring nodes tend to have similar labels and features).
 
\subsubsection{Aligning label-conditioned distributions} To effectively align label-conditioned distributions across domains, many existing works directly model conditional distribution shifts with discrepancy measurement, such as MMD and its variants~\cite{CDNE,mywork1}. Besides, DANE~\cite{DANE2} calculates class centroids from labeled nodes and works to minimize the cumulative distances between same-class centroids across domains. It helps to alleviate the negative impact of outliers as centroids are more robust to represent label-conditioned information. Furthermore, AdaGIn~\cite{AdaGIn} proposes an adversarial adaptation approach with conditional GAN~\cite{condDA}, which conditions domain discriminator on the discriminative information from source classifier predictions. CoCo~\cite{COCO}, on the other hand, employs contrastive loss on cross-domain samples within the same class to encourage similar embeddings. Moreover, considering pseudo-labels could be inevitably noisy, SA-GDA~\cite{SA-GDA} applies spectral augmentation to reduce conditional shifts in the spectral domain. 
\subsection{Task Shift}
While the majority of GDA models assume consistency in the prediction task across domains, GCONDA~\cite{GCONDA} observes for the first time that
both graph heterophily (i.e., nodes with dissimilar features and labels are connected) and GNN architecture exacerbate task shift, leading to potential performance degradation. Therefore, it explicitly models task shift with Wasserstein distance between predicted label distributions from source and target domains and reduces both task and marginal shift to achieve adaptation. 

\subsection{Multi-Channel Shift}
When the task-specific embeddings have multiple channels, a natural problem is how to effectively reduce domain shifts in multiple embedding spaces. A typical scenario is heterogeneous graphs, where various node and edge types or multiple meta-paths lead to multi-channel node embeddings. MusDAC~\cite{MusDAC} makes the first attempt in this area. It first employs a multi-channel Graph Convolution Network (GCN) encoder based on meta-paths to project nodes into multiple embedding spaces with different semantic information, then it selects a subset of embedding spaces and fuses them into a single embedding space where domain adaptation techniques could be used. 

However, the capacity of MusDAC is limited by the expressive power of encoder and the selection mechanism, which may under-explore rich semantics of heterogeneous graphs. To overcome these problems, HGA~\cite{HGA} employs powerful heterogeneous GNNs (HGNNs) to extract multi-channel node embeddings based on different meta-paths, then separately aligns the semantic-specific embeddings across domains. 
Generally, heterogeneous graphs in source and target domains contain both shared and private node types, but domain shifts caused by private node types are largely under-explored. Therefore, GDA-HIN~\cite{GDA-HIN} proposes a low-rank matrix completion method to enhance the embeddings of private nodes and separately reduces domain shifts for shared types and private types. 

In addition to studying heterogeneous graphs, MS-CNC~\cite{MS-CNC} mitigates the negative impact of long-tail nodes on adaptation by augmenting the original edges with structural and attribute similarity, resulting in three channels for each node embedding: two channels represent embeddings obtained from augmented views, while the third is their concatenation. Consequently, MS-CNC aggregates domain shifts computed from each channel. Moreover, DyTrans~\cite{DyTrans} studies the problem of dynamic graph transfer learning and reduces domain shifts in both spatial and temporal channels. 

\section{Target-based GDA}
Overall, target predictions from source learners could be inferior mainly for two reasons. Firstly, distribution mismatches across domains may still exist, rendering target embeddings inferior to the tasks. Secondly, the source learner could underfit on target domains as little or no supervision comes from them. In response, recent research has developed methods to enhance target embeddings or directly refine target predictions. They could be divided into two groups depending on the type of target information they use. We introduce their details in the following sections.

\subsection{Utilizing sample semantics}
Semantics of target samples could be leveraged to improve target embeddings, including their raw feature and the probability distributions of predicted labels. For example, TDAN~\cite{TDAN} addresses noises introduced during adaptation by reconstructing raw edge features from edge embeddings, resulting in noise-free target embeddings. Besides, minimizing the entropy of target prediction distributions could promote clear decision boundaries in target embeddings~\cite{UDAGCN,mywork1}. In addition, some studies utilize target pseudo labels as auxiliary supervision. For instance, CDA~\cite{CDA} generates reliable target pseudo labels for supervised training and computes target classification loss accordingly. In the context of graph anomaly detection, ACT~\cite{ACT} employs off-the-shelf anomaly detectors and thresholding method on target embeddings after adaptation, aiming to generate anomaly and normal nodes as pseudo labels. It then retrains a new graph encoder and classifier with these labeled samples to derive improved target predictions. 

\subsection{Utilizing topological information}
\label{sect::unsup_tar}
Topological structures in the target domain encode domain-specific knowledge, which can be used to enhance target predictions. For example, DM-GNN~\cite{DM-GNN} directly refines predictions by implementing a label-propagation node classifier, which smooths each node’s label prediction by combining its own and neighbors’ predictions. In addition to prediction refinement, some methods focus on enhancing the quality of target embeddings. DAGrade~\cite{DAGrade} enhances domain-specific information learning by reconstructing the target domain's topology. Moreover, SOGA~\cite{SFUGDA} improves discriminability by encouraging the structural consistencies between target nodes in the same class. Such consistencies include nodes being connected or having similar structural roles.

\section{Extensions}
\label{sect::ext}
Most of the existing GDA works primarily concentrate on simplified adaptation tasks with one source domain (fully accessible) and one target domain, and the task is completely consistent across domains. However, real-world tasks are more challenging, and these assumptions may not hold. In this section, we introduce those pioneering studies which address more complicated graph adaptation tasks.

\noindent\textbf{Source-Free GDA}: In source-free tasks, the primary challenge is leveraging the discriminability of the well-trained source model $P(Y^{S_i}\vert X^{S_i})$, which is the only available source knowledge with $D^{S_i}$ inaccessible. Aligning distributions between source and target domains is not feasible in this case. Therefore, SOGA~\cite{SFUGDA} maximizes the mutual information between the target graph and its corresponding predictions, enhancing the certainty of target predictions.

\noindent\textbf{Multi-Source GDA}: When there are multiple source graph domains available ($m^S>1$), each source domain varies in similarity to the target domain, and there are also inherent differences among sources. Therefore, a key problem is how to integrate and transfer the most task-relevant knowledge from source domains. NESTL~\cite{nestl} trains separate models for each source domain and linearly combines these models with topological similarity between each source-target pair. Besides, MSDS~\cite{msds} selects a fixed number of the most transferable source domains for a particular target domain. It jointly considers three types of distances between cross-domain raw feature distributions, i.e., Maximum Mean Discrepancy (MMD), Wasserstein Distance, and Jensen–Shannon divergence.

\noindent\textbf{Universal GDA}: Universal GDA assumes there are both common classes ($\mathcal{Y}^{com}=\mathcal{Y}^{S}\cap\mathcal{Y}^{T}$) and private classes ($\mathcal{Y}^S\setminus\mathcal{Y}^{com}$, $\mathcal{Y}^T\setminus\mathcal{Y}^{com}$) across domains. Source learners should correctly predict a target sample if it belongs to $\mathcal{Y}^{com}$; otherwise, mark it as `unknown'. Therefore, UDANE~\cite{UDANE} initially separates nodes belonging to common and private classes by ranking the prediction uncertainty of target nodes and selecting the nodes with highest certainties as common set. Subsequently, embeddings of target nodes in the common set are aligned to source embeddings for adaptation. As target nodes outside the common set are not aligned, their predictions may be unstable and inaccurate. Thus, UDANE employs contrastive learning to enhance the quality of target embeddings and thus improve predictions.

\section{Applications}
\label{sect::app}
\noindent\textbf{Anomaly Detection}: In graph-based anomaly detection problems, the complexity of anomaly signals makes labeling a significant challenge. Therefore, researchers have explored transferring extensive labeling knowledge from related domains to mitigate label scarcity. Two main strategies have emerged in the existing literature. Firstly, the conventional cross-entropy loss is replaced by deviation loss to highlight anomaly signals~\cite{GDN,ACT}. Besides, as certain types of anomalies in target domain may not appear in source domain, they can not directly benefit from the transferred knowledge. To enhance the qualities of overall target embeddings, Commander~\cite{COMMANDER} utilizes additional neighborhood-reconstruction scores to compute node-level anomaly scores, while ACT~\cite{ACT} constructs a group of anomaly and normal nodes as target pseudo labels and retrains the encoder and anomaly detector. DAGrade~\cite{DAGrade} alternatively designs an auxiliary task of reconstructing target topology.

\noindent\textbf{Urban Computing}: An important area of urban computing is studying modern physical systems, where the extensively distributed sensors generate a wealth of networked data for nodes and edges with complex spatial-temporal correlations. GNA~\cite{GNA} studies transferring knowledge between groups of physical systems by formulating the problem as cross-domain spatial-temporal graph classification. Specifically, it integrates the structures, temporal changes, and label information to derive a specialized graph kernel, which is further used to compute domain discrepancy with MMD. Another critical application is traffic prediction, wherein models usually rely on large-scale historical traffic data which is not always available in real-world scenarios. Thus, CrossTReS~\cite{CrossTReS} transfers knowledge in traffic data across cities with meta-learning framework. It trains graph encoders with node- and edge-level domain adaptation techniques to reduce domain shifts arising from different distributions of regions (nodes) and roads (edges). In addition, CrossTReS transfers more knowledge from source regions that are more similar to target regions, enhancing the effectiveness of adaptation.

\section{Conclusion and Prospects}
In this paper, we present a comprehensive survey on graph domain adaptation, in which we systematically categorize and analyze the existing literature based on the proposed taxonomy. We further discuss the extensions and applications. In addition to the recent progress, numerous challenges and opportunities await future exploration, and we summarize some promising prospects as follows.

\noindent\textbf{Designing tailored DA models for graphs}: As the shift in complex graph structures is one of the primary causes of graph domain divergence, it is worth investigating to develop more exclusive DA models for structured graph data by integrating specific structural attributes (e.g., subgraphs) into the GDA model design. 

\noindent\textbf{Selecting optimal source domains}: Currently, the field of GDA lacks thorough research on pre-selecting the most suitable source domains for transferring knowledge to a particular target domain, which could guarantee transfer effectiveness in advance. However, due to the complex structural distributions in graph data, calculating the relevance between source and target domains is challenging. Therefore, we need to design specialized algorithms and explore the corresponding theoretical foundations for this purpose, wherein tools and theories from graph kernels or graph signal processing could be leveraged.

\noindent\textbf{Exploring tasks with limited target data}: 
GDA typically requires access to the data distributions in target domains. Hence, in tasks with scarce or no target domain data, GDA models need to enhance the ability to learn general and foundational knowledge from source domains. One feasible direction is integrating graph pre-training techniques. Additionally, we can explore other research fields, such as domain generalization on graphs, to improve the model's generalization to unknown target domains.

\bibliographystyle{named}
\bibliography{ijcai24}

\end{document}